# Creating a New Persian Poet Based on Machine Learning

Mehdi Hosseini Moghadam      Bardia Panahbehagh

Department of Mathematics, Faculty of Mathematics and Computer Science,
Kharazmi University, Tehran, Iran [1]

Oct 16, 2018

In this article we describe an application of Machine Learning (ML) and Linguistic Modeling to generate persian poems. In fact we teach machine by reading and learning persian poems to generate fake poems in the same style of the original poems. As two well known poets we used Hafez (1310-1390) and Saadi (1210-1292) poems. First we feed the machine with Hafez poems to generate fake poems with the same style and then we feed the machine with the both Hafez and Saadi poems to generate a new style poems which is combination of these two poets styles with emotional (Hafez) and rational (Saadi) elements. This idea of combination of different styles with ML opens new gates for extending the treasure of past literature of different cultures. Results show with enough memory, processing power and time it is possible to generate reasonable good poems.

## 1 Introduction

Have a moment and think how we use our brain and intelligence to make words and sentences that have meaning and can express our ideas and insights. For example as pointed by McKeown (1992): " in the process of producing discourse, speakers and writers must decide what it is that they want to say and how to present it effectively. They are capable of disregarding information in their large body of knowledge about the world which is not specific to the task at hand and they manage to integrate pertinent information into a coherent unit. They determine how to appropriately start the discourse, how to order its elements, and how to close it. These decisions are all part of the process of deciding what to say and when to say it. Speakers and writers must also determine what words to use and how to group them into sentences. In order for a system to generate text, it, too, must be able to make these kinds of decisions." Now imagine how hard that would be to teach such ability to a computer. Recently there have been many attempts to generate texts that are both syntactically and semantically correct by computers (Misztal-Radecka and Indurkhya, 2016; Wang et al., 2017; Vinyals and Le, 2015; Leopold, 2014). With the help of artificial intelligence, computers are now able to do such difficult tasks. Also with the help of deep learning (recurrent neural network), computers can generate words and sentences that make sense (Xie et al., 2017; Sutskever et al., 2011; Graves, 2014). But first lets have a historical example of text generation. *George Philipp Harsdörffer* was a German poet in the 1600s who belonged to a literary society. He created the *Fünffacher Denckring der Teutschen Sprache* in 1651 (Figure 1). This contraption (which translates as "The Five-fold Thought-ring of the German Language") was a set of five concentric circles with letters and word fragments written on them: prefixes on one ring, starting letters on another, then middle letters, ending letters and finally suffixes. The idea was that you stacked the circles together and twisted them around independently to generate different words to act as poetic inspiration. You could leave the end-word rings in place while twisting around the start-word rings, so it acts as a kind of rhyming dictionary. You could make an existing word and then change a syllable or two to see what happened. Or you could just twist all the circles round to make new words and look at them and think about what they would mean if they were real.[2]

---
[1] panahbehagh@khu.ac.ir ; m.h.moghadam1996@gmail.com

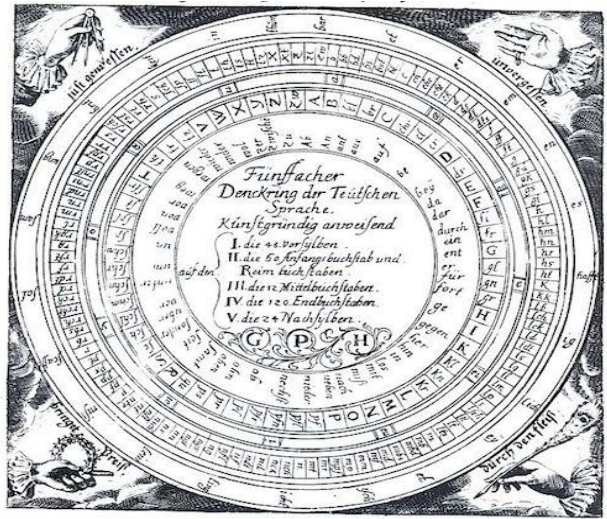

Figure 1: Fünffacher Denckring der Teutschen Sprache; set of five concentric circles with letters and word fragments written on them: prefixes on one ring, starting letters on another, then middle letters, ending letters and finally suffixes.

In the field of literature, text generation can be used as a tool for reproducing historical texts and literary legacy. As we know in the history of every culture there exist many famous and well known poets and writers with brilliant works, so the ability to generating poem and text almost based on their styles can be of a great interest.
Here our purposes are:

• By feeding our text generator model the poems of Hafez, make machine able to compose Hafez style poems.
• By feeding our text generator model the poems of Hafez and Saadi, make machine able to compose combined style poems.
For the models we use the data set of Ghazaliat-e-Hafez and Ghazaliat-e-Saadi[3].

## 2 Initial concepts

We define some concepts which may be useful for understanding the process of ML text

---

[2] http://mathesonmarcault.com/index.php/2015/12/15/randomly-generated-title-goes-here/

[3] the Mohammad Qazvini/Ghāsem Ghani 1941 edition

generation:

**Sequential data:** Is a type of data which stores in a chronological order.

**Long term dependency**: Consider this phrase 'Clouds are in the __', in order to fill in the the blank we have enough information. Just 4 words before the blank is enough to predict the blank, so we need a little information, this little dependency is called short term dependency. Now consider 'John was born in London and has passed first 5 years of his life in London … and his mother toung is__'. Now in order to fill in the blank we need more information in comparison to the first example, may be even pages of information. This long dependency to the previous words is called long term dependency.

**Artificial neural network (ANN):** Artificial neural networks (ANNs) are computing systems vaguely inspired by the biological neural networks that constitute animal brains such systems "learn" to perform tasks by considering examples, generally without being programmed with any task-specific rules.

**Recurrent neural network (RNN):** Is a class of artificial neural network where connections between nodes form a directed graph along a sequence. This allows it to exhibit dynamic temporal behavior for a time sequence. Unlike feedforward neural networks, RNNs can use their internal state (memory) to process sequences of inputs.

**Word cloud:** An image composed of words used in a particular text or subject, in which the size of each word indicates its frequency or importance.

## 3 Model and its mechanism

In this section we describe our model.

The model is an improvement of model which is presented By Jason Brownlee on August 4, 2016 in Natural Language Processing [4], with some major changes[5].

The original model had three layers including: LSTM, Dropout, Dense; which can be run on both CPU and GPU (Figure2).

The model has seven layers including: CuDNNGRU-1, Dropout, CuDNNGRU-2, Dropout, Dense-1, Dense-2, Dense-3 (Figure 3). CuDNNGRU is a Fast GRU implementation backed by CuDNN which can only be run on GPU, with the TensorFlow backend. The model how ever is deeper than the original model with five times more number of parmeters but faster than the original one.

---

[4] https://machinelearningmastery.com/text-generation-lstm-recurrent-neural-networks-python-keras/
[5] Please contact with the writers for receiving the complete code

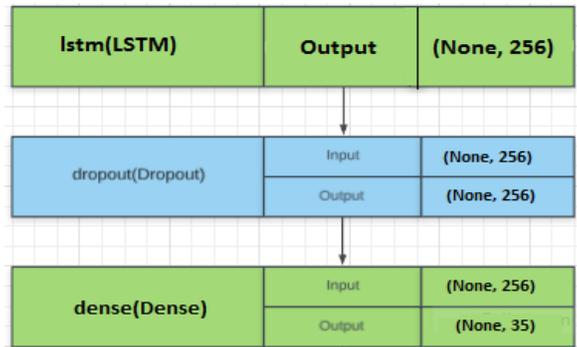

Figure 2: The original model with three layers including: LSTM, Dropout, Dense. The numbers show the shape of data which we feed in to the model.

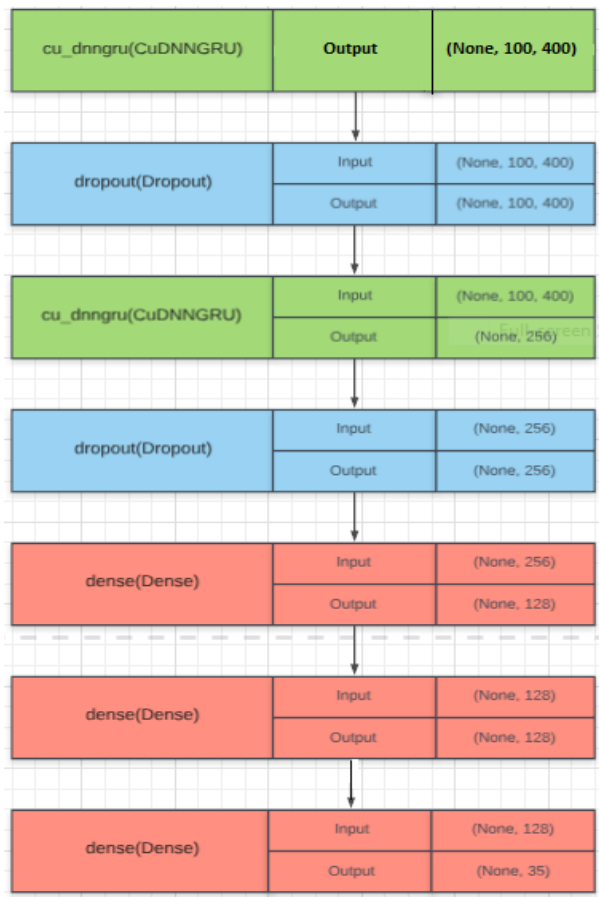

Figure 3: New model with seven layers including: CuDNNGRU-1, Dropout, CuDNNGRU-2, Dropout, Dense-1, Dense-2, Dense-3. The numbers show the shape of data which we feed in to the model.

Given any textual data to the model (in our case the poems of Hafez), after encoding, it works in three steps as follows:

• First it selects a predetermined size of characters called the first pattern. Then the second pattern is obtained by shifting the first pattern to the right with one character. For instance consider the first four patterns of the following poem[6]:

" اگر ان ترک شیرازی به دست ارد دل ما را "

"به خال هندویش بخشم سمرقند و بخارا را"

If we choose length of each pattern 20, we have the following patterns:

P1:
" به دست ارد دل ما را "

P2:
" ی به دست ارد دل ما ر "

P3:
" زی به دست ارد دل ما "

P4:
" ازی به دست ارد دل ما "

If we present the patterns in a list shape we would have the following:

P1:

| ا | ر | – | ا | م | – | ل | د | – | د | ر | ا | – | ت | س | د | – | ه | ب | – |
|---|---|---|---|---|---|---|---|---|---|---|---|---|---|---|---|---|---|---|---|

P2:

| ر | – | ا | م | – | ل | د | – | د | ر | ا | – | ت | س | د | – | ه | ب | – | ی |
|---|---|---|---|---|---|---|---|---|---|---|---|---|---|---|---|---|---|---|---|

The character " – " means blank space.
Then all these patterns will be added into a list "list of patterns". Next, the first character which comes exactly after each pattern will be added into another list "list of characters"(Figure4).

• Next, data (consisting "list of patterns" and "list of characters") feed into the model (deep network). By observing each pattern and also based on long term dependency and frequency of characters, the model memories (learns) what word come next after each pattern. For example by observing P1 and P2 it memories the words "ی" or "ز" respectively (Figure 5).

Then the model uses this information (learned from whole text and all the patterns) to predict each next word and arrange them to compose a poem or any textual data.

---

[6] Note that, Persian is written from right to left, but computer reads and processes the words from left to right.

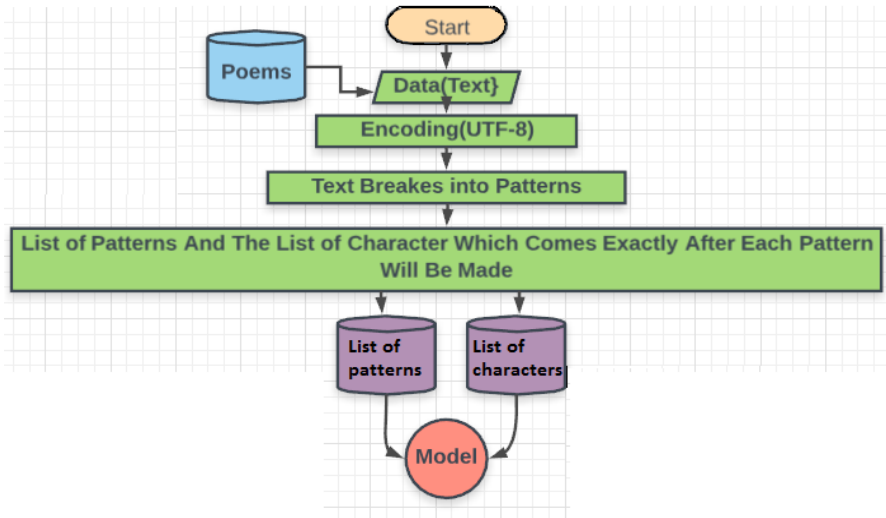

Figure 4: Textual data will be divided into patterns, then all these patterns will be added into a list ("list of patterns") and the first character which comes exactly after each pattern will be added into another list ("list of characters").

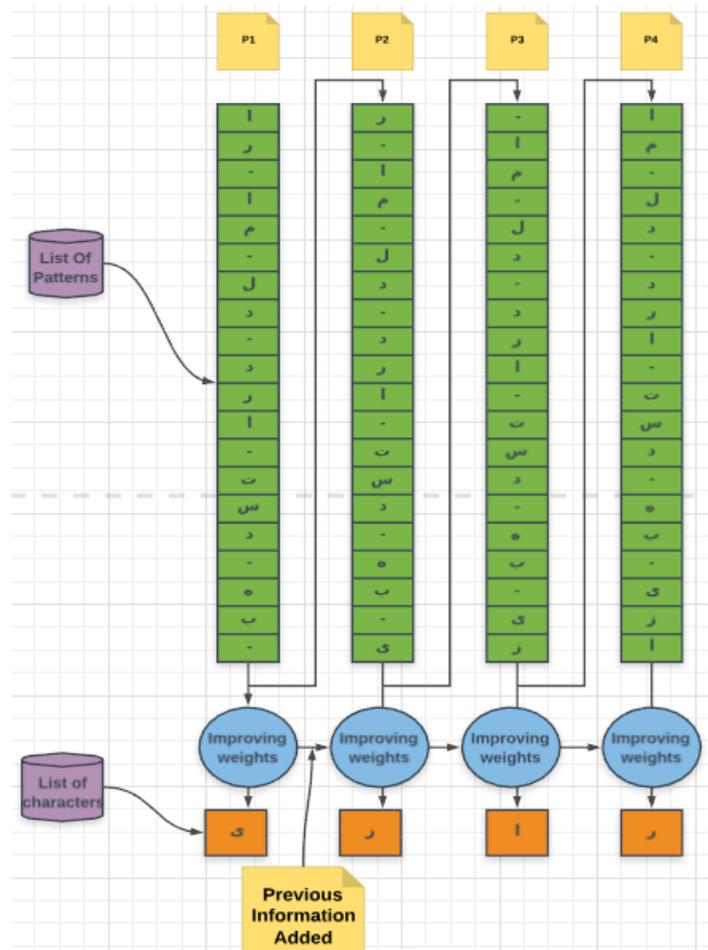

Figure 5: Deep Network; by observing each pattern and also based on long term dependency and frequency of characters, the model memories (learns) what word come next after each pattern.

Assume P1 and also S1 from

"که کس نگشود و نگشاید به حکمت این معما را"

where:

S1:

| ا | ر | - | ا | م | ع | م | - | ن | ی | ا | - | ت | م | ک | ح | - | ه | ب | - |
|---|---|---|---|---|---|---|---|---|---|---|---|---|---|---|---|---|---|---|---|

Both of S1 and P1 end with"ب ه–", and then the first case leads to predict next word as "ی" and in the second, "د". Now if the machine face to "ب ه–" it will uses all the information of epochs (long term dependency) and frequencies to predict the best characters.

• Based on our time and processing power the machine reads and processes the whole data several times, each called one epoch. Learning machines use iterative algorithms often need many epochs during their learning phase. In this step, after each epoch, the machine becomes better in predicting the characters. After processing, we give it a starting sentence (called *Seed*) and with the use of information stored in saved weights we can generate fake poems (Figure 6).

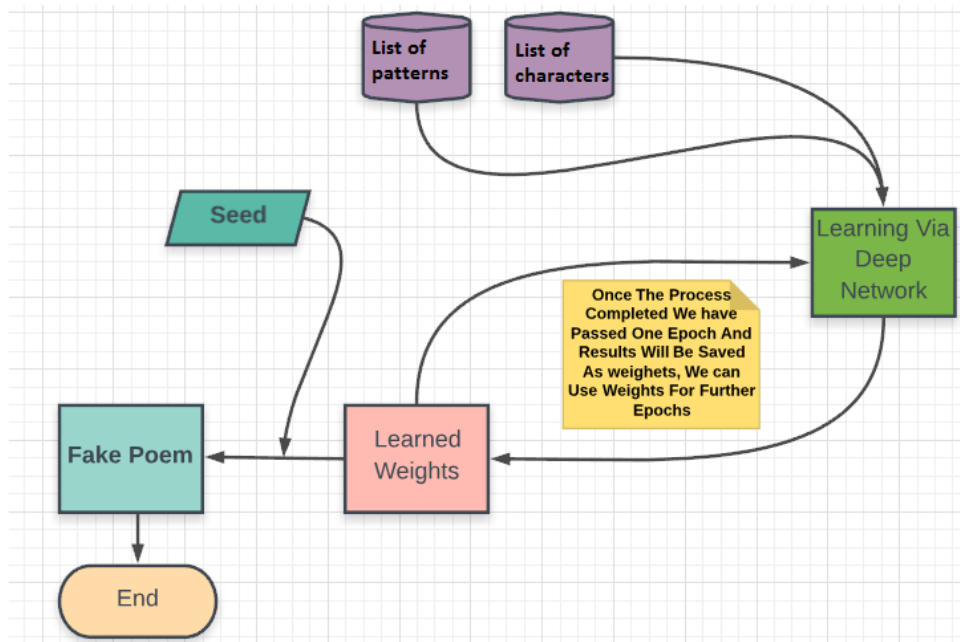

Figure 6: Once the data has been processed, we can give the model a starting sentence (called *Seed*) and with the use of information stored in saved weights we can generate fake poems.

# 4 Results

In this section we present the results of our model to generate poems in Hafez style and also a new style result of combination of Hafez and Saadi. Because of limitation of processing power, the generated text needs a little manipulation to make better sense. According to Oxford Advanced Learner's Dictionary, 8th edition, a poem is:

"a piece of writing in which the words are chosen for their sound and the images they suggest, not just for their obvious meanings. The words are arranged in separate lines, usually with a repeated rhythm, and often the lines rhyme at the end."

Our generated poems have some differencesses and similarities with this definition, as follows:

Similarities:

- In both cases The words are arranged in separate lines.

- Often the lines rhyme at the end.

Differencesses:

- Not all lines suggest obvious meanings.
- Not all lines are connected in meaning, so they don't necessarily tell a coherent story.

In addition, in our generated poems not all lines necessarily obey a particular Poetry style, however majority of them do. Also with this model we can generate as many couplets as we wish, even millions of them. Firstly we give the seed, secondly we set a *limit number* for example 200, finally model generates a poem with 200 characters as follows.

Seed:

الا یا ایها الساقی ادر کاسا و ناولها

که عشق آسان نمود اول ولی افتاد مشکل ها

به بوی نافه ای کاخر صبا

200 characters generated poem:

نیست کو معلین

دامشت خانقاه به من من از در میخانه بست

در کار ما به در این خیال ما را ببرد

ز شاه ساهد می و زلف تو را تا برسد

تا به درد تو به میخانه در آب است

که با ز من آباد به در خسابات

به خاک دل بر ت

## 4.1 A poet like Hafez

Our model, first trained on poems of Hafez. Relation between epoch and accuracy of the model to make syntactically and semantically better words is presented in (Figure 7). As mentioned before if our model observes and analyses the whole Ghazaliat-e-Hafez, it has passed one epoch. It is notable that in the first epoch, the machine starts without any background about the text, but for the second epoch the machine proceed with information of the first epoch about the whole text and so on. Also accuracy is defined as the percentage of correct prediction of characters.

The results of accuracy over epochs are interesting. As we can see in (Figure 7), in initial epochs the learning rate (improving accuracy) is fast and substantial but as the time passes this rate decreases significantly. For understanding the process better, consider a new born baby. From the first day the baby begins the process of learning to speak and putting words next to each other. Consider each epoch as a week, in the first weeks, the baby learns many words and sentences since they are all new, but as the time passes the learning process gets slower because now the baby has learned a lot of sentence and there are a few words which are new for it. In the same way our model born in the first pattern in epoch 1. In the initial epochs our model is exactly like a new born baby (learns faster) but at the end of training (epoch 150) we have a baby of 3 years old!

Our model on hafez style model contains the data (text) of whole Ghazaliat-e-hafez and it has been trained over 500 epochs.

Word clouds of the real and generated texts are in (Figure 8) and (Figure 9) which again show that frequencies of the used words in the both real and generated text are similar. Now please look at some of results with different number of epochs:   50 epochs: With small number of epochs as we can see below, the machine has learned very little and it has repeated a hemistich four times:

به می به دود به دود به دوان به باد

به می به دود به دود به دوان به باد

به می به دود به دود به دوان به باد

به می به دود به دود به دوان به باد

160 epochs: In this number of epochs, the results is better but there is no rhyme:

که تو با می کشید از سر کوین من

از می کند دل از دست می نوزیرم

به خون دل بنگر که از سر من یکنی چیست

من که بر سر موی تو با من از تو دانی

500 epochs: With 500 epochs, the results are better both in meaning and rhyme:

<div dir="rtl">
چو کار بگذرد این سر که من از این جان است

که این معشب که با تو باری کار و نشان است

برون اندر این منزل و بیابان به باد

بهار از خاک دل من ای دل که تو باد
</div>

## 4.2 A new created poet

Now it is easy to feed our model a certain type of data and generate text in the same style. For instance if we feed our model the poems of hafez the model can generate hafez style poems and if we feed the model saadi style poems then we get those of like saadi's.

Now if we feed our model both hafez and saadi poems, we can generate poems in the style of both. In other words, we can create a new poet with both hafez and saadi perspective. It can lead us to era of new poets. In order to create such new poems we feed our model both Ghazaliat-e-hafez and Ghazaliat-e-saadi. Word clouds of the real and generated texts are in (Figure 10) and (Figure 11) which again show that frequencies of the used words in the both real and generated text are similar. Some results were as following:

<div dir="rtl">
به درداب و دست از خلایت کسی

که بر من کنی با تو را در کسی

که در باز در بند بر خود نهاد

که با هوش در باز در خود نهاد

به دست از تو را بر کسی بر امید

به در بار در باد دودمن امید
</div>

As we can see, these poems are neither like Hafez, nor saadi, but their styles are very similar to the both which is very interesting.

## 5  Conclusion

Using machine learning, an idea to create a new poet based on poets that have passed away, is presented. For this purpose, we let the machine to read and process the text based on machine learning algorithms to find some probabilistic patterns for arranging the characters of the original texts. Then with giving some characters in the form of words to machine as the seed, the machine based on the probabilistic patterns can generate as many characters as we want. With this method, style of the generated text will be similar to the original one. With this idea of combination of the styles of poets, we will give our poets a new life and also our culture and history a new chance to have some new poets with some new styles not previously exist.

Two of the challenges of this method are limitation of processing power and limitation of the text size. More powerful computers in processing power and the greater size of the original text leads to generating better text in terms of rhyme and concept.

For future works, it would be interesting if we combine the style of writers instead of poets. Also it would be interesting if we feed the machine with poems of some old poets and some new born words in our literature to see if our previous poets want to compose poem using new born words, what they would compose.

Figure 7: Relation between epoch and accuracy of the model to make syntactically and semantically better words. In initial epochs the learning rate (improving accuracy) is fast and substantial but as the time passes this rate decreases significantly.

Figure 8: Word cloud of the real hafez poems.

Figure 9: Word cloud of the fake generated poems of hafez.

Figure 10: Word cloud of the real hafez and saadi poems.

Figure 11: Word cloud of the fake hafez and saadi generated poems.